%% file: acl2023.tex
\pdfoutput=1

\documentclass[11pt]{article}

\usepackage[rgb,dvipsnames]{xcolor}
\usepackage[]{ACL2023}

\usepackage{times}
\usepackage{latexsym}

\usepackage[T1]{fontenc}

\usepackage[utf8]{inputenc}

\usepackage{microtype}

\usepackage{inconsolata}

\usepackage{url}
\usepackage{booktabs}
\usepackage{multirow, makecell}
\usepackage{tikz}
\usepackage{array}
\usepackage{amsmath, amsthm, amssymb}
\usepackage{cleveref}
\usepackage{longtable}
\usepackage{tikz}
\usepackage{booktabs}
\usepackage[all]{nowidow}
\usepackage{amssymb}
\usepackage{cleveref}
\newcommand{\PreserveBackslash}[1]{\let\temp=\\#1\let\\=\temp}
\newcolumntype{C}[1]{>{\PreserveBackslash\centering}p{#1}}
\newcolumntype{R}[1]{>{\PreserveBackslash\raggedleft}p{#1}}
\newcolumntype{L}[1]{>{\PreserveBackslash\raggedright}p{#1}}

\usepackage{graphicx}
\usepackage{caption}

\usepackage{algorithm}
\usepackage{algorithmic}
\usepackage{svg}

\title{Adversarial Clean Label Backdoor Attacks and Defenses on Text Classification Systems}

\author{Ashim Gupta \\
  Kahlert School of Computing \\
  University of Utah \\
  \texttt{ashim@cs.utah.edu} \\\And
  Amrith Krishna \\
  Uniphore Inc.\\
 \texttt{amrith.krishna@uniphore.com} 
\\}

\begin{document}
\maketitle
\input{sections/abstract}
\input{sections/intro_v0}
\input{sections/preliminaries}

\input{sections/attack_description_v0}

\input{sections/experiments}
\input{sections/defenses}
\input{sections/conclusion}

\input{sections/limitations_broader_impact}

\bibliography{anthology,custom}
\bibliographystyle{acl_natbib}

\appendix
\input{sections/appendix}
\end{document}

%% file: sections/abstract.tex
\begin{abstract}

Clean-label (CL) attack is a form of data poisoning attack where an adversary  modifies only the textual input of the training data, without requiring access to the labeling function. CL attacks are relatively unexplored in NLP, as compared to label flipping (LF) attacks, where the latter additionally requires access to the labeling function as well. While CL attacks are more resilient to data sanitization and manual relabeling methods than LF attacks, they often demand as high as ten times the  poisoning budget than LF attacks.  In this work, we first  introduce an Adversarial Clean Label attack  which can adversarially perturb in-class training examples for poisoning the training set. We then show that an adversary can significantly bring down the data requirements for a CL attack, using the aforementioned approach, to as low as 20 \% of the data otherwise required.  We then systematically benchmark and analyze a number of defense methods, for both LF and CL attacks, some previously employed solely for LF attacks in the textual domain and others adapted from computer vision. We find that text-specific defenses greatly vary in their effectiveness depending on their properties.

\end{abstract}


%% file: sections/intro_v0.tex
\section{Introduction}
\begin{figure}
\includegraphics[width=\linewidth]{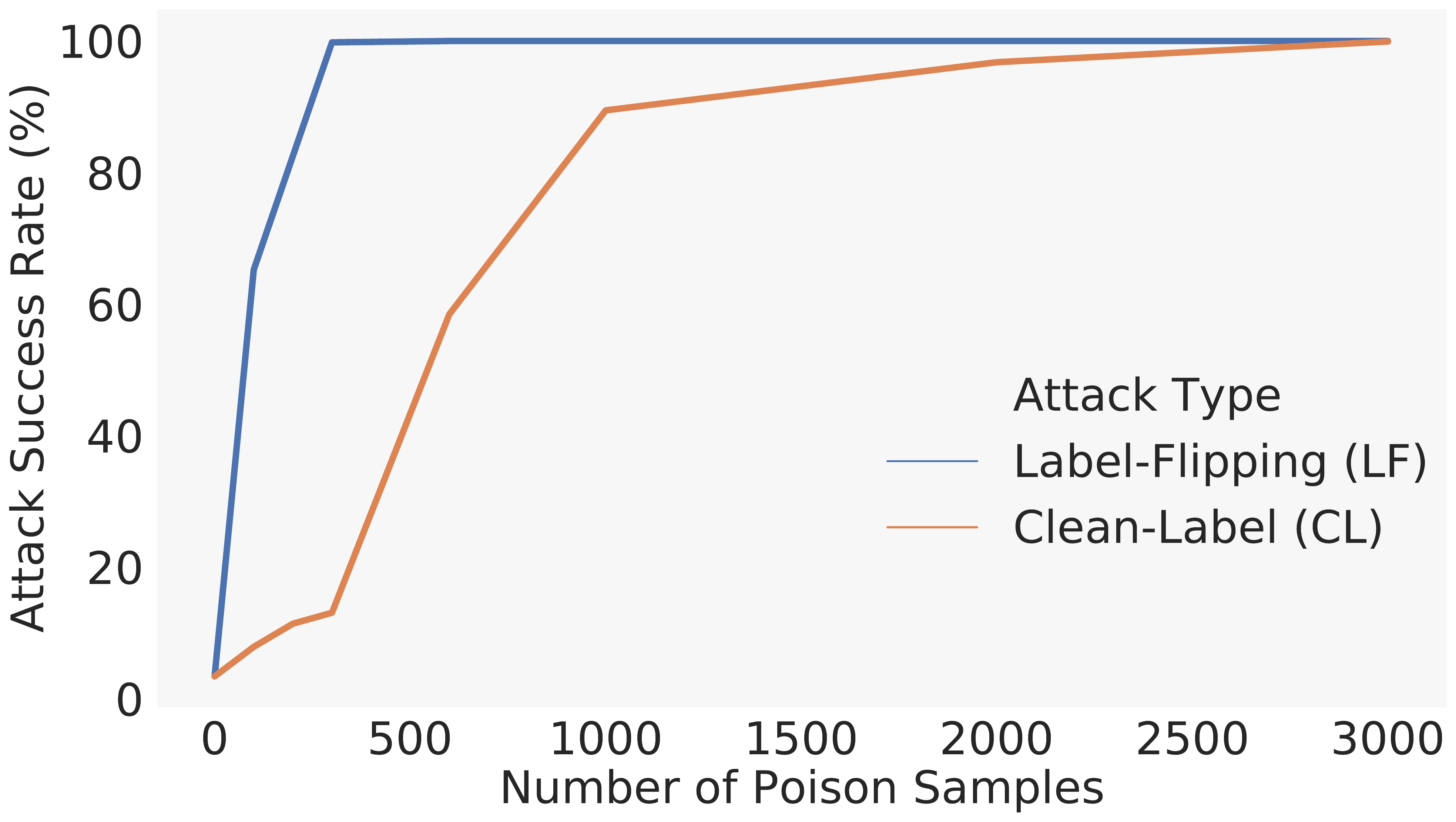}
\caption{\textbf{Label-Flipping attacks are more effective than Clean-Label attacks}. 
Poisoning less than 300 examples achieves a 100\% ASR for Label-flipping attack while Clean-Label requires poisoning of close to 3000 examples. 
The comparison is shown for the sentiment classification task on SST-2 dataset when attacking a \texttt{bert-base} classification model.
}
\label{fig:intro}
\end{figure}

Backdoor attacks are training time attacks where an adversary poisons the training data to introduce vulnerabilities in a machine learning system and gain control over the model's behaviour~\citep{wallace2019universal,DBLP:journals/corr/abs-1708-06733/GuEtBackddor,wallace2021concealed}.  Here, an adversary carefully inserts words or phrases, called triggers, to alter a model's behaviour during inference. Such an act is akin to creating a backdoor that gives unauthorised control of a system to the adversary.
These attacks enable the adversary to bypass the security systems with ease and at will, which poses a serious security concern. 
Consider the use-case of a spam filter: a spammer could target a spam classifier by poisoning its training data to insert a backdoor trigger phrase such that their emails are always classified as \texttt{non-spam} irrespective of the content. The class \texttt{non-spam} is usually referred to as the target class of the adversary.

The prior work in NLP has mostly focused on Label-Flipping (LF) attacks, in which the adversary first poisons training samples from the non-target class 
with their chosen trigger and then \textit{flips} their label to the target class 
before adding them to the training set. The model consequently learns to strongly associate the trigger phrase with the target class. Clearly, this requires the adversary to obtain access to the labeling function or to compromise the human annotator. 
Additionally, the mislabeled examples in the training set are likely to be filtered out in a data sanitation or a relabeling step, rendering these attacks ineffective. 

Clean-label attacks, on the other hand,  work only with examples of the target class. Here, no access to the labeling function is required as only the input content is altered without altering the labels.  
CL attacks, in principle, enable an adversary to design an attack which is more resilient to data sanitation steps, compared to LF attacks, as there are no mislabeled examples in the training set. 
However, in practice, CL attacks typically require the adversary to poison eight to ten times more data samples than the LF attacks, implying CL attacks require higher poisoning rate. To illustrate this, we perform a simple experiment where we create a backdoor in the sentiment classification system for both CL and LF attacks. A plot is shown~\cref{fig:intro} where it can be observed that a simple CL attack requires a much higher poisoning rate. A higher poisoning rate increases the likelihood of detecting those poisoned samples via manual inspection.


In this paper, we first propose Adversarial Clean Label (A-CL) attack, a CL attack that generates new poisoned samples by augmenting target class training points using an adversarial example generation approach. Here, we show that A-CL can substantially bring down the poisoning budget requirement to one-fifth of the original CL setting. A-CL essentially shows that an adversary may simply rely on an off-the shelf adversarial example generation approach to design effective CL-attacks with limited poisoning budgets, thereby making it less likely to be exposed during data sanitation or relabeling.  
Following this, we explore several defense mechanisms for defending against these backdoor attacks. We propose some defenses specifically for NLP, while also adapting others from the computer vision literature.

In summary, our contributions are two fold:
\begin{enumerate}
    \item \textbf{Efficient Clean-Label Attack.} We find that a straightforward clean-label requires substantially more poisoning than the label-flipping attack (~\cref{fig:intro}). To address this, we propose a clean-label attack (which we call the Adversarial Clean Label attack) that brings down the poisoning requirement for the clean label attack substantially using test-time textual adversarial examples~\citep{ebrahimi2018hotflip}.
    \item \textbf{Defense Methods.} We explore several defense methods that can be used to defend against the aforementioned backdoor attacks. Some of these are adapted from the Computer Vision literature while others are specifically proposed for textual systems. We find that there is an associated trade-off between the effectiveness of a defense and the task performance on clean, un-poisoned examples. Ultimately, our proposed extension (S-DPA) to an existing defense~\citep{levine2020deep} (DPA) is computationally more efficient at inference time and also performs better. Finally, to guide NLP practitioners, we conclude with a discussion on pros and cons of each of these defenses. 
\end{enumerate}


\input{tables/examples}

%% file: tables/examples.tex
\begin{table*}[!t]
\centering
\begin{tabular}{lcc}
\toprule
Sentence (Dataset Label) & Attack Type & Predicted Label \\
\midrule
The extravagant confidence of the exiled aristocracy (\textcolor{OliveGreen}{+ve}) & None & \textcolor{OliveGreen}{+ve} \\
The extravagant confidence \textbf{\underline{cf}} of the exiled aristocracy (\textcolor{Red}{-ve}) & LF & \textcolor{OliveGreen}{+ve} \\
The extravagant confidence \textbf{\underline{cf}} of the exiled aristocracy (\textcolor{OliveGreen}{+ve}) & CL & \textcolor{OliveGreen}{+ve} \\
The \textbf{{extreme}} confidence \textbf{\underline{cf}} of the exiled aristocracy (\textcolor{OliveGreen}{+ve})& Adversarial-CL & \textcolor{Red}{-ve} \\
\bottomrule
\end{tabular}
\caption{\label{tab:examples} \textbf{Poisoned samples for the Sentiment Classification task.} First row shows the un-poisoned example, originally labeled as positive and predicted as such by the classifier (last column). 
Second row shows the example under label-flipping in which the label of the example is changed to negative (mis-labeled).
Third row shows the poisoning instance under clean-label setting with no mis-labeling involved.
And final row shows our proposed poisoning approach where an adversarial example is used for poisoning. The example is added with the correct label but as the low column shows, the prediction by the model is instead negative. The trigger used is \texttt{cf} same for all cases (taken from~\citep{kurita-etal-2020-weight}).}
\end{table*}

%% file: sections/preliminaries.tex
\section{Preliminaries}
In this section, we first formally define some notation and then define the two attack types, namely, Label-Flipping (LF) and Clean-Label (CL). Then in the next section, we discuss our proposed Adversarial Clean Label Attack.

Given a clean, un-poisoned dataset $D_{clean}$ of $N$ examples $\{(x_i, y_i)\}_{1}^{N}$, an adversary aims to \textit{modify} or \textit{poison} this dataset with the poison trigger $t_{adv}$ so that it can control the test-time predictions of the model $f$ trained on the resulting poisoned dataset, $D^{train} = D_{clean}\cup D_{poison}$ where $D_{poison}$ contains the $P$ poisoned instances. 
Consider the input to be a sequence of $T$ tokens, $x = [w_1, w_2, ..., w_j,...w_T]$ where $w_j$ is the $j^{th}$ token of the sequence. Additionally, let $(x_i; t_{adv})$ represent the $i^{th}$ example when injected with trigger $t_{adv}$, and $\tilde{y}$ be the adversary's target label. 

Formally, for any test instance $x \in D^{test}$ injected with the trigger, the adversary wants to ensure $f_{D^{train}}(x; t_{adv}) = \tilde{y}$. Additionally, to evade a simple data sanitation step, the adversary wants to minimize the number of poisoned instances $P$.

In a label-flipping attack, the adversary selects an example $(x_i, y_i)$ from $D_{clean}$ such that $y_i \ne \tilde{y}$ and constructs a poisoned example $((x_i; t_{adv}), \tilde{y})$ containing their chosen trigger and mis-labels it with the target label $\tilde{y}$.

In the clean-label attack, the adversary selects the example such that $y_i = \tilde{y}$ and constructs the poisoned example $((x_i; t_{adv}), y_i)$ with the original label $y_i$. Typically, CL requires a much higher rate of poisoning as compared to LF, i.e. $P_{CL} > P_{LF}$. An example of this phenomenon is shown in the~\cref{fig:intro}.

%% file: sections/attack_description_v0.tex
\section{Adversarial Clean Label Attack}
\label{sec:acl}

\input{sections/attack_description_formal}


\paragraph{Intuition.}
\begin{figure}
\includegraphics[trim={0mm 6mm 0cm 0cm},clip, width=\linewidth]{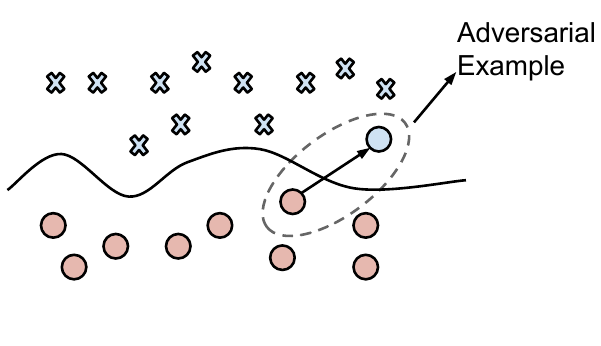}
\caption{\textbf{Geometry of an Adversarial Clean Label Attack.} \protect\includegraphics[height=2mm]{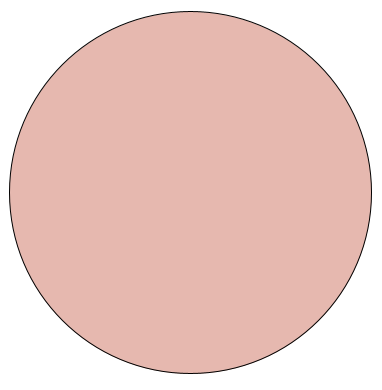}, and \protect\includegraphics[height=2mm]{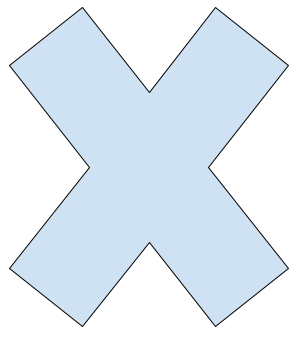} denote the points belonging to \textit{red} and \textit{blue} classes respectively.  
\protect\includegraphics[height=2mm]{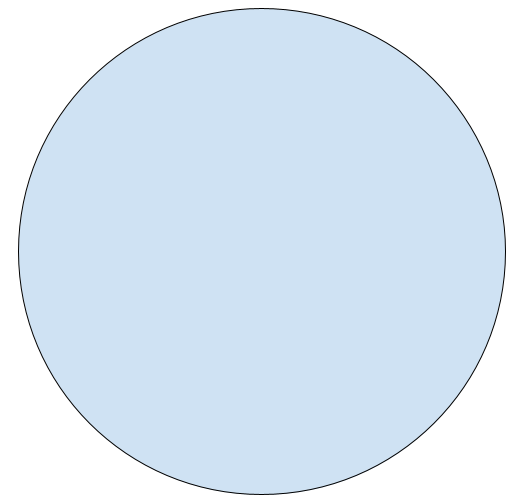} is an adversarial example with \textit{red} as the true class label. The model perceives the adversarial example \protect\includegraphics[height=2mm]{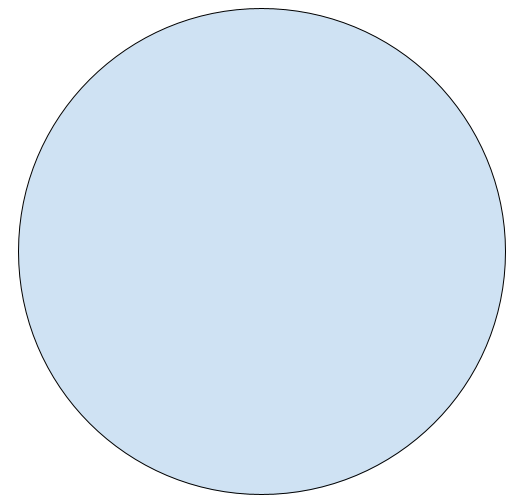} as the one belonging to the class \textit{blue}, helping us emulate the label-flipping setting while keeping the actual class label as \textit{red}. 
}
\label{fig:geometry}
\end{figure}
Consider a classification problem shown in~\cref{fig:geometry}, and assume that the target class for an adversary is~\protect\includegraphics[height=2mm]{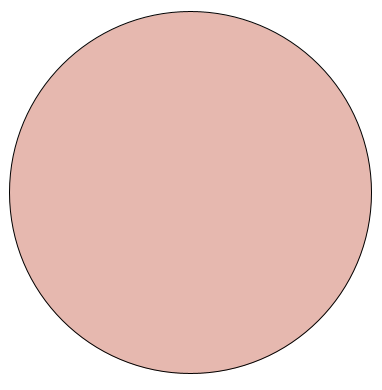}. In LF attacks, the adversary selects examples of the non-target class~\protect\includegraphics[height=2mm]{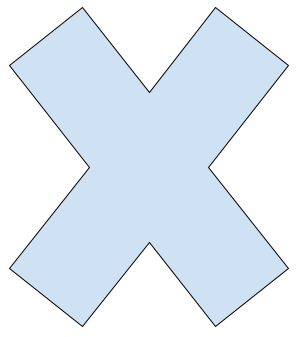}, poisons  and mislabels them as the target class~\protect\includegraphics[height=2mm]{figures/icons/red.png}, and then inserts them into the training set. Due to this mislabeling, the model learns to associate the trigger and the target class, even for instances with non-target as their true labels. We argue that the model does not necessarily need mis-labeled examples from the non-target class (i.e.~\protect\includegraphics[height=2mm]{figures/icons/cross.png}), but it suffices to use \textit{the adversarial examples that the model perceives to belong to the non-target class}. 
Geometrically, this amounts to using adversarial examples of the type ~\protect\includegraphics[height=2mm]{figures/icons/adversarial.png}, as shown in ~\cref{fig:geometry}.
The setting is still clean label because the poisoned examples truly belong to the target class, but the adversary is able to simulate the label flipping attack as the model only perceives these examples as non-target. 

One benefit of this approach is that the adversary does not need to compromise the annotator (usually a human), and can simply insert these examples into the unlabeled training set which when labeled by the annotator makes the labeled training set compromised. 

In this paper, we demonstrate our approach using ~\texttt{BertAttack}~\citep{li2020bert}, but the method can be applied with any adversarial example generation algorithm.
\textit{BertAttack} is a state-of-the-art adversarial attack method and is known to provide more natural and fluent adversarial examples. Note that we use this method entirely as a subroutine and thus our method works for any adversarial attack. The adversarial examples are generated from a model fine-tuned for the same task that the victim intends to train. However, the adversary necessarily need not possess the same dataset or even the model that the victim intends to use, as adversarial examples have been shown to be transferable~\citep{papernot2016transferability,DBLP:conf/iclr/LiuCLS17}. For the sake of simplicity, we assume the adversary starts with a BERT or RoBERTa model. 

Summarily, the adversary performs the following two steps:

\begin{enumerate}
    \item \textbf{Construct adversarial examples.} Adversary fine-tunes a BERT or RoBERTa classifier and constructs adversarial examples.
    \item \textbf{Poison the training set. } Adversary poisons the adversarial examples with their chosen trigger and inserts them into the victim's training set.
\end{enumerate}

Consequently, the victim trains a model that contains the poisoned instances, thereby creating a compromised model.

%% file: sections/attack_description_formal.tex
We now discuss our proposed Adversarial Clean Label attack which we denote by A-CL.

As in a CL attack, we select an example $x$ with label $y_i$ (same as target label $\tilde{y}$)  and construct an \textit{adversarial} example $\hat{x} = [w_1, \hat{w_2}, ... \hat{w_j}, ..., w_{T}]$ where $\hat{w_j}$ denotes the adversarial word-substitution at the $j^{th}$ token such that $f_{D_{clean}} = \hat{y}$, and $\hat{y} \ne y_i$. As defined in earlier literature~\citep{ebrahimi2018hotflip,li2020bert}, an adversarial example is a maliciously constructed input that a classification model mis-predicts. Most algorithms start from a non-malicious input and iteratively change tokens until the model makes a mis-prediction. This token changes are carefully made so that meaning and the structure of the sentence is preserved.

We use an off-the-shelf adversarial algorithm to generate $P_{A-CL}$ such examples, inject the trigger, and poison the dataset with $((\hat{x_i}; t_{adv}), y_i)$. Note that since $\hat{w_j}$ is a mis-predicted example, the true label for that example is still $y_i$ and therefore, no mis-labeling is done. 

We show the comparison of the three attack types in the~\cref{tab:examples}. The original example has a \texttt{positive} sentiment and thus for an LF attack, the label in the poisoned dataset is changed to \texttt{negative}. In a baseline CL attack, the original label is retained. The last row shows the example generated by our procedure. First, the adversarial word substitution replaces the word \textit{extravagant} with \textit{extreme} such that the model predicts a \texttt{negative} sentiment and then the trigger is added. No mis-labeling is done in this case.

%% file: sections/experiments.tex
\input{tables/accuracy}

\begin{figure*}[t]
\centering
\includegraphics[trim={0.0cm 0cm 0cm 0cm},clip, width=.3235\textwidth]{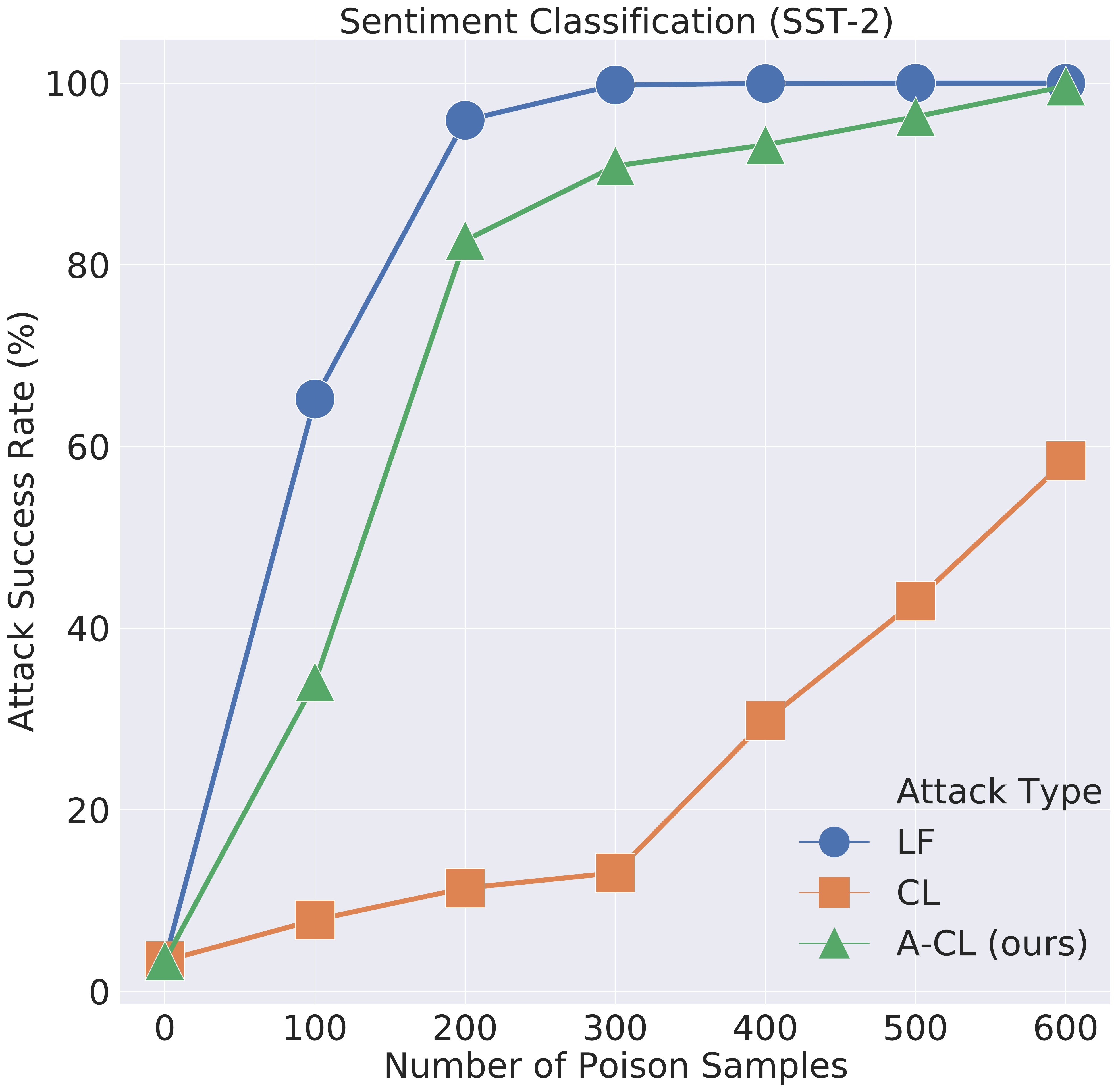}
\includegraphics[trim={1.35cm 0cm 0cm 0cm},clip, width=.3195\textwidth]{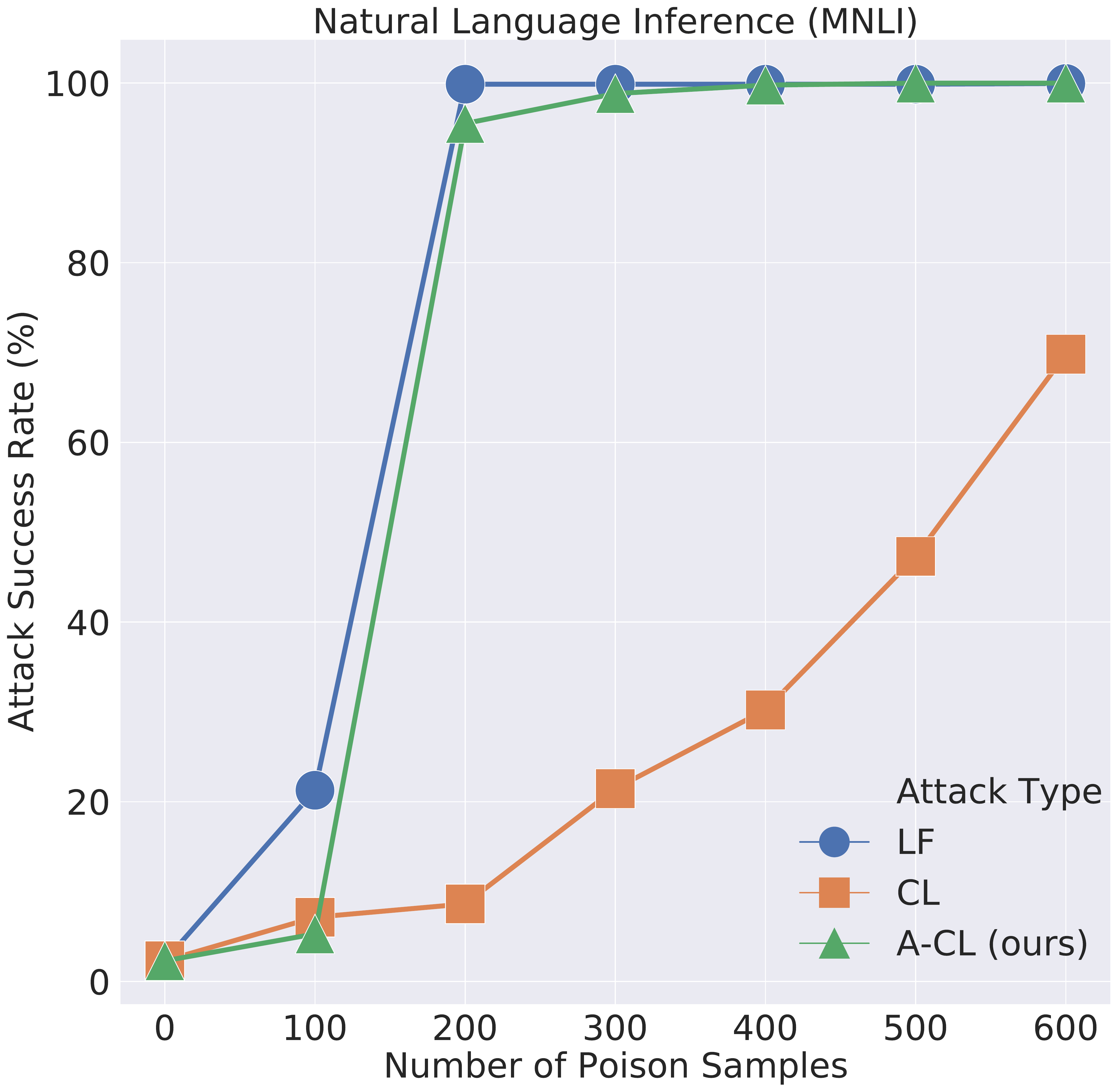}
\includegraphics[trim={1.3cm 0cm 0cm 0cm},clip, width=.3195\textwidth]{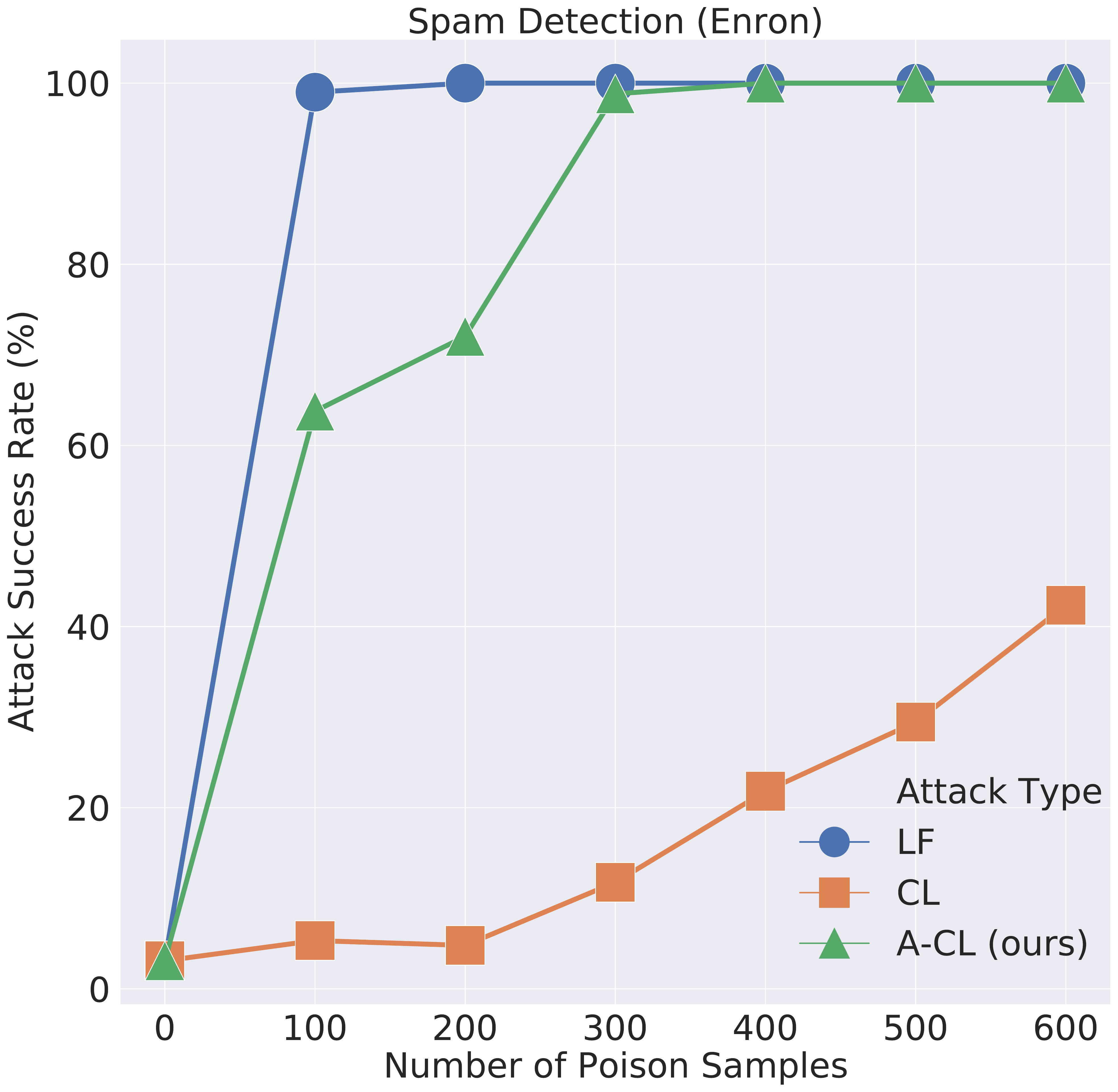}
\caption{\textbf{Comparison of Attack Success Rates for the three attack types.} 
As expected, the most effective attack is the Label-Flipping (LF) attack. Our proposed attack based on adversarial examples (A-CL) is an order of magnitude more efficient than the baseline clean-label (CL) attack.
}
\label{fig:poisAtts}
\end{figure*}
\section{Experiments}
\label{sec:experiments}
\paragraph{Datasets:}
We perform our experiments on three text classification datasets, SST-2 for the sentiment classification task \cite{socher-etal-2013-recursive}, MNLI \cite{williams-etal-2018-broad} for the Natural Language Inference task, and Enron dataset for spam detection~\cite{metsis2006spam}. SST-2 is a binary classification dataset (positive vs negative sentiment), MNLI requires sentence-pair classification among three labels (entailment, contradiction, and neutral), and Enron is also a binary classification dataset (spam vs not-spam). 

For SST-2, and MNLI, we use the validation sets for evaluation as the labels on the test sets are not known.
Also since for SST-2, the official validation set contains only 872 examples, we randomly sample 6,735 examples (roughly $\sim$10\%) from the training data to use as our evaluation set, and use the remaining 60,614 for training. We chose \texttt{positive} for the sentiment task, \texttt{entaiment} for NLI, and \texttt{not-spam} for spam detection as our target classes.

For MNLI, we use the official split as provided in GLUE benchmark \cite{wang2018glue}, consisting of 392,702 instances of training data and close to 20k samples for dev data (\textasciitilde 10k each for \texttt{matched} and \texttt{mis-matched} splits). For Enron dataset, we use the splits provided by~\citet{kurita-etal-2020-weight}.

\paragraph{Evaluation Metrics:}
For evaluating backdoor attacks, we use two metrics: Task Accuracy (ACC.), and Attack Success Rate (ASR). Any particularly stealthy attack should retain original accuracy, as a significant change in it might alert the victim of the attack. ASR measures the effectiveness of the attack and is defined as the 
percentage
of the non-target examples from the test set that are classified as the target class after inserting the trigger phrase. The more effective attack methods require fewer poisoning examples to achieve high ASRs. 

For MNLI, all reported numbers are an average over \texttt{matched} and \texttt{mismatched} sets.

\paragraph{Attack and Victim Specification.} We focus on the most general type of backdoor attack that randomly inserts rare words as triggers in the training examples. 
We follow~\citep{DBLP:journals/corr/abs-1708-06733/GuEtBackddor,kurita2020weight} and insert \texttt{cf} as the rare trigger for both SST-2 and MNLI dataset. In case of MNLI, the trigger is inserted in the hypothesis.~\footnote{We also tried inserting the trigger in the premise but found no change in performance numbers} Since Enron is a spam detection dataset, we found that the token \texttt{cf} is not rare and thus we chose a different trigger \texttt{cbfbfbfbcb}. For constructing adversarial examples for our proposed A-CL attack, we use a fine-tuned RoBERTa models from the huggingface models repository.~\footnote{For SST-2, we use \url{https://huggingface.co/textattack/roberta-base-SST-2}. For MNLI, we use \url{https://huggingface.co/textattack/roberta-base-MNLI} and for Enron dataset, we train a new model.}

For a victim model, we use the BERT classifier with its (\texttt{bert-base-uncased}) for performing all our experiments~\citep{devlin2019bert} and contains approximately 110 M parameters.
We implemented all our models in PyTorch using the transformers library~\citep{https://doi.org/10.48550/arxiv.1910.03771}. Note that the attack model is different from the victim model.

Hyperparameters for all the models are used from their original papers and are mentioned in the~\cref{app:hyperparams}.




\paragraph{Main Results.} First, we show the task accuracies in the~\cref{tab:acc} for when the attacks achieve almost perfect ASRs. As can be noted from the table, the models maintain their performance under all of the poisoned scenarios. Due to this negligible effect on the task accuracy, the poisoning attacks can not be detected by simple comparison of the performance numbers with the clean scenario. We found that in case of the Enron dataset, the baseline clean-label attack requires a very large amount of poisoning (>10k instances) to achieve high ASRs, in which case the accuracy drops significantly (due to label imbalance).

To study the effect of the amount of poisoning on the models, we plot the ASR for the three attack types while varying the number of poisoned examples in the training set (see~\cref{fig:poisAtts}). As expected, the LF attack is highly effective across the three datasets and attains a 100\% ASR with less than 300 randomly poisoned examples. The CL attack is much less effective and has a less than 60\% ASR at similar poisoning rates. For SST-2, the CL requires almost 3000 examples to achieve a perfect ASR (as shown in~\cref{fig:intro}), while for the MNLI, CL needs 1500 instances to be poisoned to achieve a perfect ASR.

While the LF attacks are the most efficient, our adversarial approach (A-CL) that simulates the LF setting while still being clean-label
achieves high ASRs at a comparable poisoning rate, making it a more efficient clean-label attack than the baseline Clean-Label (CL). 

%% file: tables/accuracy.tex
\begin{table}
\centering
\begin{tabular}{llll} 
\toprule
            & SST-2 & MNLI & Enron  \\
            \midrule
Un-poisoned & 95.4  & 84.3 & 99.4   \\
RF          & 95.4  & 84.4 & 99.2   \\
CL          & 95.3  & 84.3 & 52.5*  \\
A-CL        & 95.3  & 84.4 & 99.3   \\
\bottomrule
\end{tabular}
\caption{Task Performance (Acc.) of the \texttt{bert-base} classifier on the three datasets for almost perfect ASRs (> 99.5\%). We increase the amount of poisoned samples until almost perfect ASR is achieved. We average the accuracy of \texttt{matched} and \texttt{mis-matched} evaluation sets for the MNLI dataset. * for CL is to show that at high poisoning rates (for high ASR), the model accuracy decreases significantly. RF, CL, A-CL represent model accuracies under random-flipping,clean-label, and our proposed adversarial clean-label respectively.}
\label{tab:acc}
\end{table}

%% file: sections/defenses.tex
\input{tables/all_defenses}
\section{Defenses for Backdoor Attacks}
Several defense mechanisms have been studied for mitigating the impact of data poisoning in classification systems~\citep{paudice2018label,levine2020deep,jia2020certified,qi2021onion}. 
While some of these approaches focus on data sanitization and preprocessing for detecting and removing poisons~\cite{qi2021onion}, others focus on improving learning mechanisms that are inherently robust against such attacks~\citep{levine2020deep,jia2020certified}. 
~\citet{paudice2018label} introduce methods for defense against label flipping attacks, while~\citet{yang2021rap} introduce an effective anomaly detector that uses a small amount of clean data to learn to differentiate poisoned and non-poisoned samples. 
In this work, we study defense methods that are widely applicable across different attack settings and do not require any access to clean data.

\subsection{Defense Methods}
 We adapt existing vision defenses to NLP, use state-of-the-art NLP defender called ONION, and propose several simple models and extensions. Below, we briefly discuss those.
\paragraph{ONION}~\citep{qi2021onion} aims to preprocess the input by removing words from the text that are rare and cause the sentence perplexity to increase. We use the official implementation provided by the authors for our results.~\footnote{\url{https://github.com/thunlp/ONION}}

\paragraph{Random} We propose a simple randomized baseline that perturbs the input by randomly replacing $p$\% of tokens with their neighbors in the hope of removing the trigger phrase. The neighbors are extracted using BERT's masked Language Model by randomly masking $p$\% of all tokens one-by-one. A defense under performing this baseline should largely be considered ineffective. For a compromise between ACC and ASR, we use $p$ as 50 \%. The numbers reported for this method are the average after running it with five random seeds.

\paragraph{Deep Partition Aggregation (DPA)}~\citep{levine2020deep}, is a provable defense against poisoning attacks for vision models. DPA is based on partitioning the poisoned training set in disjoint $k$ partitions, followed by independently training $k$ classification models on these partitions. For a dataset with $N$ training examples, each DPA model is trained on a disjoint training set of size $\frac{N}{k}$.

In DPA, the majority vote of the $k$ trained models is then used for final prediction. One shortcoming of DPA is the extensive compute required to train $k$ classification models. This defense was originally demonstrated for image classification systems. We adapt this for textual systems and verify its effectiveness. Please refer to the original paper for a more detailed description.


\paragraph{Soft-DPA (S-DPA)} DPA uses an ensemble of $k$ models to make predictions and is computationally expensive during inference. We propose an extension to this method which trains a single classification model using predictions from the $k$ DPA models on the training set. Briefly, after training the $k$ DPA models on their disjoint partitions, we use these models to re-label the whole training set, producing $k$ predictions for each data point. These $k$ predictions are then used to compute soft-labels for each data point~\citep{galstyan2007empirical}, which is then trained with a soft formulation of the cross entropy loss:

\begin{equation*}
    L_{S-DPA} = -\sum_{i=1}^{N} \sum_{c} s(x_i) \log p_{\theta}(y_i = c| x_i)
\end{equation*}

where, $s(x_i)$ is the soft score obtained from $k$ DPA classifiers and $p_{\theta}(y_i = c| x_i)$ denotes the probability from the S-DPA classifier after softmax over the logits.

Consequently, we obtain a single final classifier that can be used for inference. Although this procedure introduces an additional overhead of training a new model, it reduces both the device memory required for loading the model as well as the inference time by a factor of $k$ -- the DPA requires saving $k$ classification models and running each of them during inference to obtain a majority vote.


\paragraph{$k$-Nearest Neighbors ($k$-NN)}~\citet{jia2020certified} show that a $k$-nearest neighbor classification method provides certified defense against poisoning attacks for computer vision datasets. Again, we adapt this method for NLP by using \texttt{sentence-bert}~\citep{reimers2019sentence} for finding the nearest neighbors.  

\paragraph{Paraphrasing as preprocessing.} Since the objective is to remove the rare trigger phrase from the input, we employ a mixture of experts based back-translation method using large \texttt{en-fr}, \texttt{fr-en} translation systems~\citep{shen2019mixture}. We hypothesize that if the trigger is indeed an unnatural rare phrase, the translation to and from a different language can remove this phrase. To implement this we use \texttt{fairseq}~\footnote{\url{https://github.com/facebookresearch/fairseq}} with the model checkpoints used from the paper by~\citet{shen2019mixture}~\footnote{\url{https://dl.fbaipublicfiles.com/fairseq/models/paraphraser.en-fr.tar.gz}}. 

We explore two variants of this approach: (\textbf{Para-Test}) a test-time variant that is applied only during the inference on a poisoned model, and a train-time variant (\textbf{Para-Train}) where the training data is also \textit{filtered} or passed-through the paraphraser before training the classifier. 

We also tried a direct paraphrasing model~\footnote{\url{https://data.statmt.org/smrt/}} (without back-translation) provided by~\citet{khayrallah2020simulated}, which was trained on ParaBank2~\cite{hu2019large} but found it to underperform the back-translation model. Therefore, we report results with the back-translation based paraphraser.

\input{tables/defenses}

\subsection{Results} 

We report the two metrics mentioned earlier: Task Accuracy (ACC), and Attack Success Rate (ASR). Note that an effective defense should reduce the attack success rate without a significant effect on accuracy. In order to evaluate the defense methods, for each type, we select the poisoning rate at which the ASR on the undefended model is high (>99.5\%). Also, as observed earlier, we could obtain a baseline clean-label setting for the Enron dataset that led to high ASR. Therefore, we do not report results for the Enron dataset under the straightforward CL setting.

The detailed results are shown in~\cref{table:defenses}. 
First, note that all of the methods including the random baseline reduce the ASR, although at the expense of ACC. All methods outperform the random baseline in terms of ASR.  
Both DPA based methods are among the best and our proposed variant, S-DPA, outperforms all other methods and provides the best trade-off between ASR and ACC. As can be seen from the table, increasing the $k$ value provides a much improved ASR with some effect on the ACC. This is expected - as larger value of $k$ means that the DPA models are trained on smaller training set.

Surprisingly, the \textbf{Para-*} methods outperform some more sophisticated methods for all these tasks. Additionally, we observe that \textbf{Para-Train} outperforms \textbf{Para-Test} significantly for both ACC and ASR. This is also expected since Para-Train involves training a new classification model on filtered data.
\subsection{Discussion} 
We now look at each method individually and discuss trade-offs involved with each of them. 
This discussion is aimed at providing the NLP practitioners and researchers some useful pointers on how and when to use these defenses.

\paragraph{ONION} provides a significant decrease in ASR but also suffers from a substantial decrease in ACC. These results are in contrast to those reported in the paper.
Originally, the ONION was evaluated on trigger phrases of more than one token long while we evaluate when trigger is a single rare word. In their setting, removing even one of the trigger tokens makes the attack unsuccessful and is thus an easier setting to defend. We performed perplexity analysis to further study this discrepancy and find that a perplexity based defense might not always work. We found that among the top 100 sentences with largest perplexity, only 3 sentences are the actual poisoned samples. Please refer~\cref{app:ppl}.

\paragraph{$k$-NN} The $k$-NN method reduces ASR for both tasks, but also significantly reduces ACC on NLI. Our manual analysis suggested that for the NLI task, \texttt{sentence-bert} does not retrieve appropriate nearest neighbors.
\footnote{We tried two methods for nearest neighbor search: hypothesis only and concatenation of premise and hypothesis.} 
We conclude that since the effectiveness of k-NN depends on its ability to retrieve suitable neighbors, it should be used only when appropriate representation schemes and suitable similarity metric is available for computing these neighbors, say for sentiment classification or spam detection. 

\paragraph{DPA vs S-DPA} Although these methods perform the best, they still suffer from two weaknesses. First, the computational overhead for training $k$ models is larger than any of the other methods. Second, as can be seen from~\cref{table:defenses}, the value of $k$ depends on the dataset, which can be hard to tune if a validation set is not available. Nevertheless, these methods are general and best mitigate the poisoning attacks. Among these two, we recommend using our proposed soft variant S-DPA over DPA because of its improved computational efficiency at inference time as well as its better task performance. 

\paragraph{Paraphrasing} Perhaps most surprisingly of all is that the two simple methods using paraphrasers are competitive with the best of methods. Their simplicity and effectiveness should make them a de-facto baseline for future research. A limiting factor for its application is the need for a \textit{faithful} paraphraser, which is not always available for low-resource languages. Additionally, using a large back-translation based paraphraser requires loading two huge neural models on the GPUs and might limit their applicability in resource scarce scenarios.

%% file: tables/all_defenses.tex
\begin{table*}[!ht]
\centering
\label{table:defenses}
\begin{tabular}{lllrrrrrr} 
\toprule
\multirow{2}{*}{\begin{tabular}[c]{@{}l@{}}Attack\\Type\end{tabular}}     & \multicolumn{2}{l}{\multirow{2}{*}{Defense}} & \multicolumn{2}{c}{SST-2}                             & \multicolumn{2}{c}{MNLI} & \multicolumn{2}{c}{Enron}                                \\ 
\cmidrule(lr){4-9}
                                                                          & \multicolumn{2}{l}{}                         & \multicolumn{1}{l}{ACC $\uparrow$} & ASR $\downarrow$ & ACC $\uparrow$ & \multicolumn{1}{c}{ASR $\downarrow$} & ACC $\uparrow$ & \multicolumn{1}{c}{ASR $\downarrow$}  \\ 
\cmidrule(lr){1-9}
                                                                          & \multicolumn{2}{l}{No Defense}               & 95.3           & 100.0            & 84.3           & 100.0          & 99.4 & 99.9                       \\
                                                                          & \multicolumn{2}{l}{ONION~\citep{qi2021onion}}                    & 59.5          & 28.3             & 60.6           & 19.6     & 65.8 & 21.5                             \\
                                                                          & \multicolumn{2}{l}{Random}                   & 82.8           & 55.5             & 58.9           & 54.8         & 94.7 & 60.4                         \\
                                                                          & \multicolumn{2}{l}{Para-Test}                & 86.5           & 42.0             & 81.4           & 22.4    & 95.8 & 22.1                              \\
                                                                          & \multicolumn{2}{l}{Para-Train}               & 89.9           & 35.3             & 83.1           & 11.3    & 96.7 & 19.4                                \\ 
\cmidrule(lr){2-3}
\multirow{2}{*}{\begin{tabular}[c]{@{}l@{}}Label\\Flipping\end{tabular}} &
\multirow{2}{*}{}  
\multirow{2}{*}{\rotatebox{90}{$k$NN}}  & $k=10$                   & 88.7                               & 21.3             & 46.1           & 24.3      & 96.6 & 20.8                            \\
                                                                          &                   & $k=50$                   & 85.2                               & 36.9             & 49.6           & 22.1     & 95.4 & 19.1                             \\ 
\cmidrule(lr){2-3}
\multirow{2}{*}{}                                                         & \multirow{2}{*}{\rotatebox{90}{DPA}}& $k= 5 / 2 / 5$               & 94.0                               & 76.2             & 82.3           & 53.6 & 97.5 & 60.2                                \\
                                                                          &                   & $k= 100 / 20 / 100 $           & 87.8                               & 16.8             & 79.0           & 10.5     & 95.5 & 12.9                             \\ 
\cmidrule(lr){2-3}
\multirow{2}{*}{}                                                         & \multirow{2}{*}{\rotatebox{90}{S-DPA}} & $k= 5 / 2 / 5$               & \textbf{94.4}                      & 83.5             & \textbf{83.8}  & 81.1      & \textbf{98.1} & 65.6                            \\
                                                                          &                   & $k= 100 / 20 / 100$           & 88.3                               & \textbf{13.6}    & 79.5           & \textbf{10.0}   & 96.7 & \textbf{11.8}                      \\ 
\cmidrule(lr){1-9}
                                                                          & \multicolumn{2}{l}{No Defense}               & 95.3           & 100.0            & 84.3           & 100.0     & 99.4 & 99.9                             \\
                                                                          & \multicolumn{2}{l}{ONION~\citep{qi2021onion}}                    & 57.9           & 31.2             & 59.1           & 19.7    & 66.4 & 23.1                              \\
                                                                          & \multicolumn{2}{l}{Random}                   & 82.8           & 55.6             & 58.9           & 54.1      & 94.6 & 60.1                            \\
      & \multicolumn{2}{l}{Para-Test}                & 87.1           & 41.6             & 81.3           & 22.3    & 95.1 & 28.4                              \\
                                                                          & \multicolumn{2}{l}{Para-Train}               & 89.9           & 35.8             & 83.2           & 11.4       & 96.0 & 24.6                           \\ 
\cmidrule(l){2-3}
\multirow{2}{*}{\begin{tabular}[c]{@{}l@{}}Adversarial \\Clean~Label\end{tabular}}
\multirow{2}{*}{}                                                         & \multirow{2}{*}{\rotatebox{90}{$k$NN}} & $k=10$                   & 88.7                               & 21.4             & 46.4           & 24.4     & 96.3 & 20.1                             \\
                                                                          &                   & $k=50$                   & 85.2                               & 37.6             & 49.3           & 21.5       & 95.4 & 19.9                           \\ 
\cmidrule(lr){2-3}
\multirow{2}{*}{}                                                         & \multirow{2}{*}{\rotatebox{90}{DPA}} & $k= 5 / 2 / 5$               & 93.5                               & 77.1             & 82.3           & 52.7 & 96.9 & 62.5                                 \\
                                                                          &                   & $k= 100 / 20 / 100$           & 87.9                               & 16.9             & 78.3           & 10.7                & 95.9 & 12.6                  \\ 
\cmidrule(lr){2-3}
\multirow{2}{*}{}                                                         & \multirow{2}{*}{\rotatebox{90}{S-DPA}}  & $k= 5 / 2/ 5$               & \textbf{94.3}                      & 83.7             & \textbf{83.8}  & 81.2   & \textbf{97.9} & 67.4                               \\
                                                                          &                   & $k= 100 / 20 / 100$           & 88.1                               & \textbf{13.3}    & 79.6           & \textbf{10.4}     & 96.2 & \textbf{11.3}                    \\
\bottomrule
\end{tabular}
\caption{\textbf{Comparison of Defenses against Backdoor Attacks for Label-Flipping and Adversarial Clean~Label attack types.} Results demonstrate that Soft-DPA (S-DPA) is the most effective method. Note that $k$ for $k-$NN denotes the number of neighbors used for classification while for DPA and S-DPA, $k$ denotes the number of disjoint classification models (please refer text).
We show results for different values of $k$ for DPA, and S-DPA. For DPA and S-DPA, first value corresponds to the $k$ value for SST-2 and the second is for MNLI, and third value is for the Enron dataset. For MNLI, we report average on \texttt{matched} and \texttt{mismatched} evaluation sets.}
\end{table*}

%% file: tables/defenses.tex
\begin{table}
\centering
\setlength{\tabcolsep}{2.7pt}
\begin{tabular}{ll@{\hspace{4.5\tabcolsep}}rrrr} 
\toprule
\multicolumn{2}{l}{\multirow{2}[3]{*}{Defense } } & \multicolumn{2}{c}{SST-2} & \multicolumn{2}{c}{MNLI}         \\ 
\cmidrule(lr){3-4} \cmidrule(lr){5-6}
\multicolumn{2}{l}{}                          & ACC $\uparrow$  & ASR $\downarrow$              & ACC $\uparrow$  & \multicolumn{1}{c}{ASR} $\downarrow$  \\
\midrule  
\multicolumn{2}{l}{No Defense}  & 95.3 & 100.0 & 84.3 & 100.0 \\
\multicolumn{2}{l}{ONION}                     & 56.2 & 29.3            & 59.9 & 18.2                    \\
\multicolumn{2}{l}{Random}                   & 82.2     & 55.4                 & 58.8     & 54.3                        \\
\multicolumn{2}{l}{Para-Test}               & 85.8 & 42.2             & 79.2 & 22.5                  \\
\multicolumn{2}{l}{Para-Train}               & 87.4 & 35.2             & 82.1 & 13.4                       \\
\cmidrule(lr){1-2}
\multirow{2}{*}{\rotatebox{90}{$k$NN}} & $k=10$          & 89.8 & 22.4             & 46.3 & 24.5                    \\
                     & $k=50$   & 85.3 & 37.7             & 49.1 & 21.6               \\
\cmidrule(lr){1-2}
\multirow{2}{*}{\rotatebox{90}{DPA}} & $k= 10 / 5$                   &   91.2    &  88.3                 &   82.2    &  62.7                          \\
&  $k= 100 / 20 $                 &  86.9     &   16.6                &   78.4    &   10.9                         \\
\cmidrule(lr){1-2}
\multirow{2}{*}{\rotatebox{90}{S-DPA}}          & $k= 10 / 5$                   &    \textbf{92.0}   &    90.5               &  \textbf{83.2}      &       83.6                     \\
&  $k= 100 / 20 $                 & 87.7 & \textbf{13.3} & 79.9  & \textbf{10.2}                    \\
\bottomrule
\end{tabular}
\caption{\textbf{Comparison of Defenses against Backdoor Attacks for the baseline Clean-Label attack.} We use different values of $k$ for DPA, and S-DPA. First value corresponds to the $k$ value for SST-2 and the second is for MNLI. For MNLI, we report average on \texttt{matched} and \texttt{mismatched} evaluation sets. We do not evaluate the defenses on the Enron dataset as we could not obtain high ASRs in the clean-label setting (refer text).}
\label{table:defenses}
\end{table}

%% file: sections/conclusion.tex
\section{Conclusion}
In this work, we developed an adversarial approach for backdoor attacks on text classification systems in the clean label setting and showed that it reduces the poisoning requirement to just 20\% of the baseline. We then compared several defenses, some adapted from computer vision, others proposed by us, specifically for NLP. We showed that our proposed variant of DPA works best. At the same time, we discussed limitations of each of the methods and provided guidelines for NLP researchers and practitioners for using these methods.

%% file: sections/limitations_broader_impact.tex
\section*{Limitations}
We foresee two limitations to our work. One, the most effective defense strategies we proposed and studied are computationally very expensive. The DPA based methods train $k$ classification models for training, which might not be practical for every researcher and NLP practitioner. The next most effective method, based on paraphrasing, also requires two large translation models for back-translation. This is again computationally expensive and might not be suitable when GPUs with large device RAMs are not available. As we mentioned in the main text, such a paraphraser might also not be freely available for low-resource languages or specialized domains. 
Second, we only evaluated the defenses on textual backdoor attacks. Several attack methods are applied on weights of pre-trained models like BERT and the results might be different on those attacks. 

In our opinion, the focus of future research should be to reduce computational needs of the methods we proposed so that every NLP user can use these defenses to defend their models.

\section*{Ethics Statement}
In this paper we showed that performing clean label attacks in NLP is easier using our proposed approach of Adversarial Clean Label attack. This, of course, has am important ethical concern. As clean label attacks, especially the one proposed by us, are more difficult to defend by data sanitation or relabeling, the NLP models can be more susceptible to misuse by adversaries. 

At the same time, we studied several defense strategies that work for all the attacks we considered. Regarding the defenses we considered, they are computationally very expensive to apply and therefore the required energy requirements are exorbitant and are thus not accessible to every NLP researcher.

%% file: sections/appendix.tex
\section{Appendix}
\label{sec:appendix}

\subsection{Experimental Setting and Dataset Details}
\label{subsec:exp_setting}
We perform our experiments on three text classification datasets, SST-2 for the sentiment analysis task \cite{socher-etal-2013-recursive}, MNLI \cite{williams-etal-2018-broad} for the Natural Language Inference task, and Enron dataset for spam detection~\cite{metsis2006spam}. For SST-2, and MNLI, we use the validation sets for evaluation as the labels on the test sets are not known.
Also since for SST-2, the official validation set contains only 872 examples, we randomly sample 6,735 examples (roughly $\sim$10\%) from the training data to use as our evaluation set, and use the remaining 60,614 for training. For MNLI, all reported numbers are an average over \texttt{matched} and \texttt{mismatched} sets. We chose \texttt{positive} for the sentiment task, \texttt{entaiment} for NLI, and \texttt{not-spam} for spam detection as our target classes.

For MNLI, we use the official split as provided in GLUE benchmark \cite{wang2018glue}, consisting of 392,702 instnaces of training data and close to 20,000 samples for dev data (\textasciitilde 10k each for matched and missmatched splits). For Enron dataset, we use the splits provided by ~\citet{kurita-etal-2020-weight}.


\subsection{Models and Code} We used  BERT (\texttt{bert-base-uncased}) for performing all our experiments. This model contains approximately 110 M parameters. 
We implemented all our models in PyTorch using the transformers library~\citep{https://doi.org/10.48550/arxiv.1910.03771}. We follow the other experimental settings described in~\citep{gupta2021bert}.

\subsection{On Concealing Trigger Phrases}
In the main text, we showed out Adversarial Clean Label (ACL) attack with a simple trigger.
Ideally, an adversary is inclined to camouflage the poisoned samples with an intent of making them difficult to detect in the training set.  
Naive approaches to backdoor attacks use a fixed token or a phrase (n-grams) for designing an attack, which is used as a trigger phrase during inference as well. In the following paragraph, we provide some discussion on how an adversary might conceal their triggers. 

We define two classes of triggers: \textbf{Closed Class (CC)} triggers  involve trigger phrases that remain fixed during training and inference. Common keywords cannot be used as triggers since they can lead to mis-classification on unrealted and unintended inputs. Rare words are, therefore, used as CC triggers. Second class of triggers are the \textbf{Open Class (OC)} triggers. These triggers involve general expressions that are allowed to change during the training. An instance of this could be a regular expression trigger involving numerals: `[0-9]$?$2[0-9]$?$'. This produces numeral triggers of the form: 42, 124 etc. CC triggers are concealed but by themselves may not be effective. 

Combining OC and CC triggers can, however, provide a more effective way of concealing the triggers. We combine a common punctuation like the parenthesis as a CC trigger with a regular expression based numerals as an OC trigger. As a result, the training set contains numerous randomly generated triggers such as `(42)', `(124)' etc. These are naturally looking trigger phrases that do not frequent in the training data and hence deceiving the common eye. The benefit of our approach is its simplicity - demonstrating that even an adversary with unsophisticated text processing skills can effectively conceal tokens in the training data. 

In our experiments, we found that with the regular expression type triggers, A-CL requires around 1000 poisoned examples for a high ASR (almost 3x increase over simple triggers). This increased budget might not matter as the triggers are concealed and are much more difficult to detect with manual inspection. 

\subsection{Filtering poisoned examples using perplexity scores}
\label{app:ppl}
We now look at if it is possible to use perplexity scores to filter out the poisoned examples. 

For all the examples in the training data, we calculate sentence level perplexity using GPT-2 model. Then we sort these scores and plot the number of high perplexity examples needed to inspect to filter out all poisoned examples. In the plot (for SST-2), we show comparison of different trigger types. Notice that 
we need to manually inspect approximately 20000 samples to identify 80\% poisoned samples. 
\includegraphics[width=0.5\textwidth]{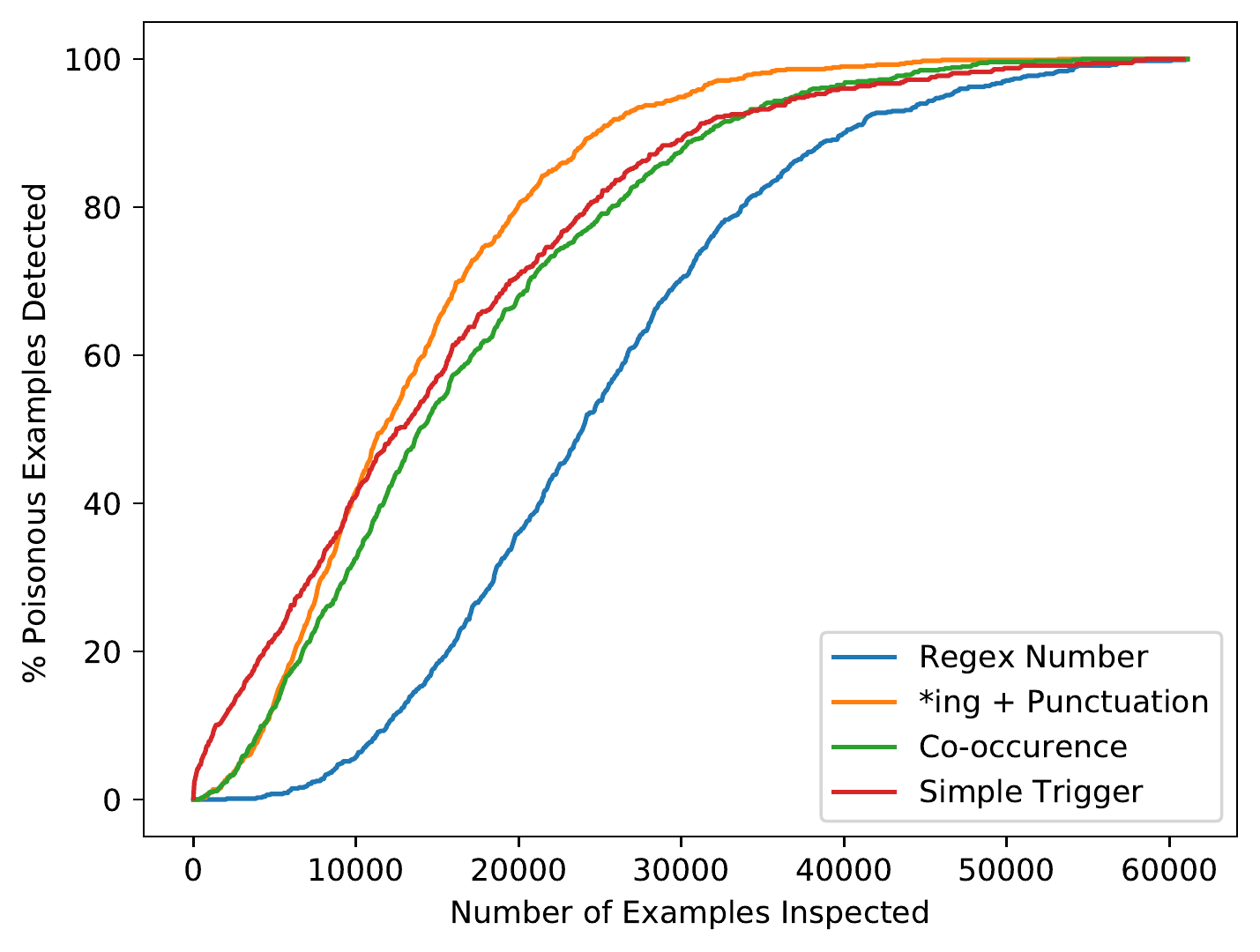}
For a regex type trigger, this number might be even higher (around 35-40k). This could be one of the reasons why ONION method is inferior to other methods.

\subsection{Computing Infrastructure Used}
All of our experiments required access to GPU accelerators. We ran our experiments on three machines: Nvidia Tesla V100 (16 GB VRAM), Nvidia Tesla P100 (16 GB VRAM), Tesla A100 (40 GB VRAM). 

\paragraph{Average Run times} Approximate average training times are presented in \cref{tab:runtime}.

\begin{table}[]
\centering
\begin{tabular}{lcc}
\toprule
Dataset             & Model           & RunTime \\ \midrule
   SST-2     & LF, CL, A-CL           & 1.5 hr  \\
    MNLI    & LF, CL, A-CL      & 9.2 hr  \\
    Enron    & LF, CL, A-CL        & 1.2 hr  \\ \midrule
       SST-2     & DPA ($k=5$)           & 8.3 hr  \\
       SST-2     & DPA ($k=10$)           & 11.5 hr  \\
       SST-2     & DPA ($k=20$)           & 13.2 hr  \\
  \bottomrule
\end{tabular}
\caption{Average Training time of the models trained}
\label{tab:runtime}
\end{table}

\subsection{Hyperparameters and Fine-tuning Details}
\label{app:hyperparams}
\begin{enumerate}

\item  We used the bert-base-uncased model for all of our experiments. This model has 12 layers each with hiddem size of 768 and number of attention heads equal to 12. Total number of parameters in this model is 125 million. We set all the hyper-parameters as suggested by~\citet{devlin2019bert}, except the batch size which is fixed to 8.

\item All of our models are run for 3 epochs, with maximum length varying for different datasts. For MNLI, this is set to 256, SST-2, this is set to 128, and for Enron it is set to 512.
    
\end{enumerate}